\documentclass[journal]{IEEEtran}
\ifCLASSOPTIONcompsoc
\usepackage[nocompress]{cite}
\else
\usepackage{cite}
\fi
\ifCLASSINFOpdf
\else
\fi
\usepackage{times}
\usepackage{epsfig}
\usepackage{graphicx}
\usepackage{amsmath}
\usepackage{amssymb}
\usepackage{multirow}
\usepackage{array}
\usepackage{bm}
\usepackage{tabularx}
\usepackage{amsmath}
\usepackage{amssymb}
\usepackage{latexsym}
\usepackage{algpseudocode}
\usepackage{tabularx}
\usepackage{booktabs}
\usepackage{url}

\usepackage{algorithmicx}
\usepackage{algorithm}
\usepackage{graphicx}
\usepackage{amssymb,subfigure,cases}
\usepackage{times}
\usepackage{color}
\usepackage{multirow}
\usepackage{algpseudocode}
\usepackage{amsmath}
\usepackage{algorithmicx}
\usepackage{algorithm}
\usepackage{graphicx}
\usepackage{tabularx}
\usepackage{amssymb,subfigure,cases}
\usepackage{times}
\usepackage{color}
\usepackage{multirow}
\usepackage{array}

\begin{document}
	
\title{AudioVisual Video Summarization}
	
	\author{Bin~Zhao,
			Maoguo~Gong, and
	Xuelong~Li,~\IEEEmembership{Fellow,~IEEE}

	\IEEEcompsocitemizethanks{
		
		\IEEEcompsocthanksitem Bin Zhao is with the Academy of Advanced Interdisciplinary Research, Xidian University, Xi’an 710071, and School of Artificial Intelligence, Optics and Electronics (iOPEN), Northwestern Polytechnical University, Xi'an 710072, P.R. China (binzhao111@gmail.com).
		\IEEEcompsocthanksitem Maoguo Gong is with the Key Laboratory of Intelligent Perception and Image Understanding of Ministry of Education, International Research Center for Intelligent Perception and Computation, Xidian University, Xi’an 710071, China.
		\IEEEcompsocthanksitem Xuelong Li is with School of Artificial Intelligence, Optics and Electronics (iOPEN), Northwestern Polytechnical University, Xi'an 710072, P.R. China.

		
	}
	
}
\IEEEtitleabstractindextext{%
\begin{abstract}
	
Audio and vision are two main modalities in video data. Multimodal learning, especially for audiovisual learning, has drawn considerable attention recently, which can boost the performance of various computer vision tasks. However, in video summarization, existing approaches just exploit the visual information while neglect the audio information. In this paper, we argue that the audio modality can assist vision modality to better understand the video content and structure, and further benefit the summarization process. Motivated by this, we propose to jointly exploit the audio and visual information for the video summarization task, and develop an AudioVisual Recurrent Network (AVRN) to achieve this. Specifically, the proposed AVRN can be separated into three parts: 1) the two-stream LSTM is utilized to encode the audio and visual feature sequentially by capturing their temporal dependency. 2) the audiovisual fusion LSTM is employed to fuse the two modalities by exploring the latent consistency between them. 3) the self-attention video encoder is adopted to capture the global dependency in the video. Finally, the fused audiovisual information, and the integrated temporal and global dependencies are jointly used to predict the video summary. Practically, the experimental results on the two benchmarks, \emph{i.e.,} SumMe and TVsum, have demonstrated the effectiveness of each part, and the superiority of AVRN compared to those approaches just exploiting visual information for video summarization.
\end{abstract}

}


\maketitle
\IEEEdisplaynontitleabstractindextext

\IEEEpeerreviewmaketitle

\ifCLASSOPTIONcompsoc
\IEEEraisesectionheading{\section{Introduction}\label{sec:introduction}}
\else

\section{Introduction}\label{section1}

Video summarization is a typical computer vision task developed for video analysis. It can distill the video information effectively by extracting several key-frames or key-shots to display the video content \cite{ji2020deep,zhao2019property}. Under the help of video summary, the viewer can perceive the information without watching the whole video. Therefore, video summary provides an efficient way for video browsing. Moreover, by removing the redundant and meaningless video content, it has potential applications in video retrieval, storage and indexing \cite{jin2020deep,li2017multiview}, as well as boosting the performance of related video analysis tasks, such as video captioning \cite{DBLP:conf/eccv/ChenWZH18}, action recognition \cite{zhang2020few}, \emph{etc..}

In the data explosion era, video summarization draws increasing attention. Lots of approaches are proposed in recent years\cite{zhu2020dsnet,apostolidis2020ac,zhao2020tth}. Existing approaches mainly summarize videos in two aspects. On the one hand, comprehensive models are developed to summarize videos according to manual criteria \cite{DBLP:conf/eccv/PotapovDHS14,DBLP:conf/icip/AnirudhMT16,lu2013story}, including representativeness, importance, interestingness and so on. They are developed to select the key-frames or key-shots that represent the whole video content, contain the important objects, have less redundancy, \emph{etc.}, so as to distill the video information effectively. On the other hand, the video data are taken as a sequence of frames.
The summary is generated according to the temporal dependencies among frames. To achieve this, the most popular recurrent neural network, Long-Short Term Memory (LSTM) \cite{hochreiter1997long}, is utilized as the backbone in video summarization. Recently, various of sequence models are developed based on it, such as Bidirectional LSTM \cite{DBLP:conf/eccv/ZhangCSG16}, Hierarchical LSTM \cite{DBLP:conf/mm/ZhaoLL17,DBLP:conf/cvpr/ZhaoLL18}, Attention-based LSTM \cite{DBLP:conf/mmm/CasasK19,fajtl2018summarizing}, \emph{etc.}. By taking advantages of the deep learning and sequential modeling ability of LSTM, they surpass the traditional approaches developed based on manual criteria, and take the leading position.

\subsection{Motivation and Overview}
Video data is naturally composed of two modalities, \emph{i.e.}, audio and vision. They record the activities from different aspects, and cooperate together to help the viewer understand the video content. Recently, multimodal learning has proved that audio and vision modalities share a consistency space, and there are semantic relations between them \cite{zhu2020deep,afouras2020self}. Lots of relevant video analysis tasks have demonstrated that the performance is promoted by utilizing the multimodal information in previous single modality tasks \cite{DBLP:conf/cvpr/Xiong0G020,li2017locality,wang2020deep}.
 
However, researchers in video summarization have not recognized the potential contributions of audio information to the performance. Most of them just consider the vision modality, and extract shallow or deep visual features to represent video frames, such as HoG (Histogram of Gradient) \cite{dalal2005histograms}, GoogLeNet \cite{DBLP:conf/cvpr/SzegedyLJSRAEVR15}, VGGNet \cite{simonyan2014very} and so on, while the audio features are ignored. 

In this paper, we argue that the audio modality can assist the vision modality to better understand the video content and structure. Concretely, the audio and vision are complementary to present activities in different modalities. For example, the music at the party reflects the pleasant atmosphere of the scene, and the cheers in soccer games indicate a good goal. However, the audiovisual inconsistency situations usually occur in videos as well. For example, the sounding object is not in the field of view. It will also bring interferences for vision modality, which is the main challenge in audiovisual video summarization.

Facing the above opportunities and challenges, we propose an AudioVisual Recurrent Network (AVRN) to jointly utilize audio and visual information in the video summarization task. To guarantee the consistency, the fusion of audio and visual information are in two-stages. The two-stream LSTM is utilized in the first stage to encode the audio and visual features sequentially, and capture their temporal dependency. Then, the audiovisual fusion LSTM is developed to exploit the consistency space between audio and visual information, and fuse them with an adaptive gating mechanism. Besides, considering that there are multi-hop storyline in the video stream due to montage and edit, the temporal neighborhood dependency are not capable enough for video summarization. In this case, a self-attention video encoder is adopted to encode the global video information. Finally, the temporal and global dependencies of audiovisual information captured by the sequence encoder and global encoder are jointly utilized for predicting the video summary.

\subsection{Novelties and Contributions}

The novelties and contributions of our work are:

1) The audio information is introduced to the video summarization task, so as to complement with the visual information in video content and structure modeling. 

2) An audiovisual recurrent network is developed to exploit the latent consistency space between the two modalities, where the mutual benefit is achieved.

3) A self-attention video encoder is conducted to capture the global dependencies among audiovisual information, which can complement with the temporal dependencies.

\subsection{Organization}

The organization of the rest paper is presented as follows. The relevant works in the literature are analyzed in Section \ref{section2}. The proposed audiovisual recurrent network is described in Section \ref{section3}. The experimental results are discussed and compared with the state-of-the-arts in Section \ref{section4}. In the final, the conclusion of our work is drawn in Section \ref{section5}.

\section{Related Works}\label{section2}

Video summarization is a long-standing computer vision task. There is a stable development in the literature. In this section, the related works are simply classified into traditional approaches and deep learning based approaches. They are reviewed in the following successively.

Earlier works devote to find a set of representatives to summarize the video content. Hand-crafted feature extractors are firstly utilized to extract feature vectors for each frame, such as color histogram \cite{li2018key}, optical flow \cite{DBLP:journals/tip/LiZL17}, histograms of gradient \cite{DBLP:conf/eccv/GygliGRG14} \emph{etc.}. To determine the representativeness, clustering algorithms and dictionary learning methods are employed to generate the video summary \cite{DBLP:journals/tip/CongLSYLL17,DBLP:conf/icip/ZhuangRHM98}. For example, \(k\)-means \cite{DBLP:journals/prl/AvilaLLA11} and \(k\)-medoids \cite{hadi2006video} allocate frames into different clusters and obtain the cluster center. Naturally, the cluster centers are viewed as the representatives and selected into the summary. Dictionary learning is also an effective method to select the representatives. It takes the frame sequence as a dictionary and tries to determine a subset of the elements to represent the original video \cite{DBLP:conf/cvpr/ElhamifarSV12,mei2015video}. Sparsity is a widely utilized prior for dictionary based video summarization. To achieve this, the \(l_0\) and \(l_{0,1}\) norms are added as the regularizer, and the block sparsity constraint is designed to speed up the convergence \cite{DBLP:conf/icmcs/MeiGWHHF14,DBLP:journals/tmm/CongYL12,ma2020video}. 

Later, researchers have realized that the representativeness is not enough to quantify the summary quality, and more manual criteria are proposed. Diversity is designed to reduce the redundancy in the summary, where similar clips are removed from the video. In this case, the key point is to determine the similarity among frames and shots. In \cite{DBLP:journals/jvcir/RenJF10}, Segeral distance is employed as the similarity metric. Furthermore, several distance metrics are combined together in \cite{DBLP:journals/jvcir/EjazTB12}, and the Determinantal Point Process (DPP) model is modified to select the diverse key-frames sequentially \cite{DBLP:conf/nips/GongCGS14}. Importance is developed to constrain the summary to maintain important objects in the video, in which, several local features are employed to determine the importance of different objects, including distance to the frame centroid, frequency of occurrence, aesthetics metrics, \emph{etc.} \cite{DBLP:journals/ijcv/LeeG15,DBLP:conf/cvpr/LeeGG12}. To model the summary comprehensively, several criteria are combined together in \cite{DBLP:journals/tip/LiZL17,tschiatschek2014learning}, including importance, representativeness, uniformity and storyness. They measure the summary quality in different aspects, and provide a comprehensive score function. 

\begin{figure*}[t]
	\centering
	\includegraphics[width=0.98\textwidth]{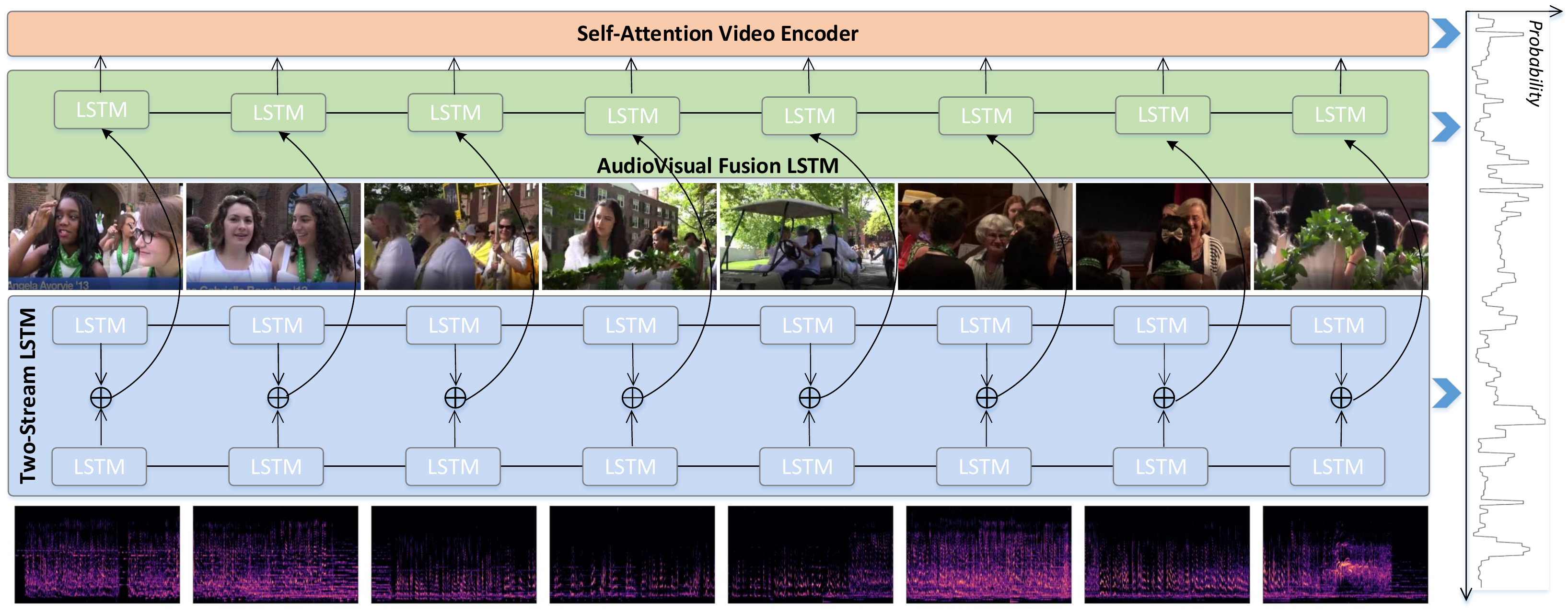}
	\caption{The architecture of the proposed audiovisual recurrent network, which is composed by the two-stream LSTM, the audiovisual fusion LSTM and the self-attention video encoder. The last row displays the log-mel spectrograms of audio data.}\label{fig2}
\end{figure*}

Recently, deep learning based approaches have made tremendous progress and taken the leading position in video summarization \cite{ji2020deep,zhao2020tth}.  Different from traditional approaches that model the summary properties explicitly with manual criteria, deep learning based approaches prefer to learn the complex mapping from video to summary directly by taking advantages of the learning ability of deep networks. Firstly, Convolutional Neural Networks (CNN) take the place of hand-crafted features to extract deep features for video frames, such as VGGNet \cite{simonyan2014very}, GoogLeNet \cite{DBLP:conf/cvpr/SzegedyLJSRAEVR15}. Then, RNNs are employed to model the frame sequence. In \cite{DBLP:conf/eccv/ZhangCSG16}, the bidirectional LSTM is developed to exploit the temporal dependency. It is further modified by adding the DPP model reduce the redundancy of the key-shot set. Considering the most favorable video length for LSTM is less than 100 frames, hierarchical LSTM is developed to extend the capability of dealing with long sequence \cite{DBLP:conf/mm/ZhaoLL17}. Moreover, a structure-adaptive LSTM is developed in \cite{DBLP:conf/cvpr/ZhaoLL18} based on the hierarchical architecture. It can promote the performance of shot segmentation, so that the independence and integrity of key-shots are preserved. 

To boost the performance, different mechanisms are developed to optimize the sequence model \cite{zhu2020dsnet,apostolidis2020ac,gao2020unsupervised}. In SUM-GAN \cite{DBLP:conf/cvpr/MahasseniLT17}, a discriminator is conducted afer the summary generator, and the generated summary is discriminated by the adversarial loss. The encoder-decoder architecture is designed in \cite{DBLP:conf/eccv/ZhangGS18}, where the decoder is utilized to recover the video content from the obtained summary in the semantic space. It can guide the encoder to select key-shots that best represent the video content. Similarly, a dual learning framework is developed in \cite{zhao2019property} based on a summary generator and a video reconstructor. In this case, the unsupervised optimization of the summary generator is achieved. The manual criteria is also adopted in \cite{DBLP:conf/aaai/ZhouQX18}, and the reinforcement learning strategy is adopted to reward the summary generator. 
\section{The Proposed Approach}\label{section3}

In this paper, we propose an AudioVisual Recurrent Network (AVRN) to integrate the audio and visual information together to the video summarization task. As depicted in Fig. \ref{fig2}, the proposed AVRN is composed of three parts, where the two-stream LSTM encodes the audio and visual feature sequentially, the fusion LSTMs fuses the audiovisual multimodal information dynamically, and the self-attention module captures the video information globally. In the following subsections, they are elaborated successively.

\subsection{Two-Stream LSTM}

LSTM is a typical recurrent neural network developed to deal with sequence data. Videos are naturally temporal sequence data. To encode the audio and visual information, the video data is separated into the audio signal and visual frames, and a two-stream structure of LSTM is conducted. 

Specifically, given the frame sequence, the visual features are firstly extracted as \({{\bf X}^v} = \left\{ {{\bf x}_1^v,{\bf x}_2^v, \ldots ,{\bf x}_n^v} \right\}\), where \(n\) stands for the length of the frame sequence. Then, to capture the temporal dependencies among frames, the bidirectional LSTM is employed as one stream to process \({{\bf X}^v}\) sequentially, which is formulated as
\begin{equation}{\bf h}_t^v = {\rm BiLSTM}\left( {{\bf x}_t^v,{\bf h}_{t - 1}^v} \right),\end{equation}
where \({\bf h}_t^v\) is the hidden state of the bidirectional LSTM. Practically, \(\rm BiLSTM\) is conducted by combining two LSTMs together. They capture the temporal dependency among frames from the forward and reverse directions, respectively, and encode the dependency into the hidden state vector \({\bf h}_t^v\) in each step.

Similarly,  to exploit the temporal dependency of audio features \({{\bf X}^a} = \left\{ {{\bf x}_1^a,{\bf x}_2^a, \ldots ,{\bf x}_n^a} \right\}\), another stream of bidirectional LSTM is conducted, \emph{i.e.},
\begin{equation}{\bf h}_t^a = {\rm BiLSTM}\left( {{\bf x}_t^a,{\bf h}_{t - 1}^a} \right).\end{equation}

In the two-stream LSTM, the temporal dependencies among audio and visual modalities are encoded into \({\bf h}_t^a\) and \({\bf h}_t^v\) separately. They are not fused correctly, and the difference and consistency between them are not considered, which may cause interference to the video summarization task. To address this problem, an audiovisual fusion LSTM is further developed. 

\subsection{AudioVisual Fusion LSTM}

In this part, an audiovisual fusion LSTM is conducted to explore the sharing latent space among the audio and visual modality, so as to exploit the consistency and reduce the difference. Different from the two-stream structure, the audiovisual fusion LSTM takes the combined audio and visual information as input and fuse them sequentially. 

Specifically, an adaptive gating mechanism is adopted to the audiovisual fusion LSTM, which is formulated as 
\begin{equation}{c_t} = {\rm Sigmoid}\left( {{{\bf W}^a}{\bf h}_t^a + {{\bf W}^v}{\bf h}_t^v + b} \right),\end{equation}
where \({{\bf W}^a}\), \({{\bf W}^v}\) and \(b\) are the training parameters. \(c_t\) is the gate to control the information flow of different modalities, which is operated as follows,
\begin{equation}{\bf x}_t^{av} = {c_t}{\bf h}_t^a + (1 - {c_t}) {\bf h}_t^v.\end{equation}
Then, the fused audio and visual information \({\bf x}_t^{av}\) is input to the audiovisual fusion LSTM to encode them sequentially,  \emph{i.e.}
\begin{equation}{\bf{h}}_t^{av} = {\rm{BiLSTM}}\left( {{\bf{x}}_t^{av},{\bf{h}}_{t - 1}^{av}} \right).\end{equation}
Practically, \({\bf{h}}_t^{av}\) captures the temporal dependencies of the audiovisual information at the \(t\)-th step, which is essential for summary generation. 

\subsection{Self-attention Video Encoder}

Although videos are naturally sequential data, multi-hops of storyline usually occur in the video stream, and the activities recorded in each shots vary largely. In some occasions, there are no obvious temporal dependencies among consecutive shots, such as those videos with edit and montage. Therefore, the sequence networks are not enough for modeling the complex video structure and content.

In this case, we develop a global encoder to cooperate with the sequence model. It is achieved by a self-attention module, 
\begin{equation} {\bf V}_t = \sum\limits_{i = 1}^n {{\alpha^i_t}} {\bf x}_i^{av},\end{equation}
where \({\bf V}_t\) is the encoded global dependency among frames at the \(t\)-th step. \(\alpha^i_t\) is the attention weight, which is computed by
\begin{equation}l_t^i = \left\langle {{{\bf{W}}^1}{\bf{x}}_i^{av},{{\bf{W}}^2}{\bf{x}}_t^{av}} \right\rangle,\end{equation}

\begin{equation}{\alpha^i _t} = \exp \left\{ {{l^i_t}} \right\}/\sum\limits_{i = 1}^n {\exp \left\{ {{l^i_t}} \right\}}, \end{equation}
where \({\bf W}^1\) and \({\bf W}^2\) are training weights. \(\left\langle { \cdot,\cdot } \right\rangle \) denotes the inner-product operation. \(l^i_t\) captures the dependency between \({\bf{x}}_i^{av}\) and \({\bf{x}}_t^{av}\). Eqn. (8) is utilized to normalize the attention weight.

\subsection{Summary Generation}

Given the computed temporal and global dependency of audiovisual information, the importance of each frame to the video content is computed by 
\begin{equation}{p_t} = {\rm Sigmoid}\left( {{{\bf W}^p}\left[ {{\bf x}_t^{av},{\bf h}_t^{av}, {\bf V}_t} \right] + {b^p}} \right),\end{equation}
where \({\bf W}^p\) and \(b^p\) are the training weight and bias. \({\bf x}_t^{av}\), \({\bf h}_t^{av}\) and \({\bf V}_t\) denote the fused audiovisual information, the temporal dependency and the global video information, respectively. They are integrated together to predict the frame-level importance. Then, the shot-level importance is obtained by averaging the importance scores of those frames in the shot, \emph{i.e.,}
\begin{equation}p_i^s = \sum\limits_{t = {s_{i - 1}} + 1}^{t = {s_i}} {{p_t}}, \end{equation}
where \( {\bf S} = \left[ {0,{s_1},{s_2}, \ldots ,{s_{m - 1}},n} \right]\) are the shot boundaries of \(m\) shots in the video. Following existing protocols, they are computed by the Kernel Temporal Segmentation (KTS) method \cite{DBLP:conf/eccv/PotapovDHS14}. In the final, the video summary is generated with those higher-score shots.

Practically, the proposed AVRN is trained end-to-end under the supervision from human-created summaries. The Mean Square Error (MSE) is utilized for optimization, \emph{i.e.,}
\begin{equation}loss = \frac{1}{n}\left\| {{\bf p} - {\bf g}} \right\|_2^2,\end{equation}
where \(\bf p\) is the predicted frame-level importance vector, \(\bf g\) is the ground truth annotated by human beings.

The proposed AVRN is operated on Python 3.7 with the deep learning platform of PyTorch 1.6. The dimensionalities of the hidden states in both the two-stream LSTM and audiovisual fusion LSTM are fixed as 256. The optimizer is selected as Adam with the learning rate 1e-5, the decay rate 0.1, decay step 30. Generally, AVRN can reach the convergence in less than 60 epochs.

\section{Experiments}\label{section4}
The experiments are carried out on two benchmark datasets, SumMe \cite{DBLP:conf/eccv/GygliGRG14} and TVsum \cite{song2015tvsum}. In the following subsections, the ablation studies are conducted, and several state-of-the-art approaches are compared.

\begin{table*}[htp]
	\centering
	\caption{The results of ablation studies for AVRN.}\label{Table1}
	\renewcommand\arraystretch{1.2}
	
	\begin{tabularx}{0.9\textwidth}{p{4cm}<{\centering}|X<{\centering}X<{\centering}X<{\centering}|X<{\centering}X<{\centering}X<{\centering}}
		\toprule
		\hline
		Datasets &\multicolumn{3}{c}{SumMe}&\multicolumn{3}{c}{TVsum}\\\hline
		Metrics &Precision  & Recall  &F-measure&Precision & Recall &F-measure  \\\hline
		Audio LSTM  &0.339 &0.380&0.354&0.538 & 0.540 &0.539  \\
		Visual LSTM  &0.376 &0.411&0.386&0.554 & 0.553 &0.554  \\
		TS-LSTM  &0.383&0.419&0.394&0.577 & 0.578 &0.578  \\
		
		AVF-LSTM   &0.389&0.417&0.396&0.567&0.568&0.567 \\
		AVRN(w/o SAVE) &0.395	&0.439	&0.415	&0.583	&0.583	&0.583\\
		AVRN(single) &0.414	&0.435	&0.419	&0.589	&0.587	&0.589\\
		AVRN &0.428	&0.464	&0.441	&0.598	&0.598	&0.597\\
		\hline\bottomrule
	\end{tabularx}
	
\end{table*}

\subsection{Setup}
\subsubsection{Dataset Introduction} 
SumMe and TVsum are popular video summarization datasets. SumMe dataset consists of 25 videos with diverse topics, including cooking, traveling, sports, \emph{etc.}. Most of them are raw videos without human edit. For each video, 15-18 users are employed to select key-shots and generate the video summaries. TVsum dataset is composed of 50 videos in open domain. They are edited videos about news, cooking, pets, \emph{etc.}. Each of them contains 20 annotations of shot-level importance scores, where the shots are generated by segmenting the video into 2-second clips evenly. 

Following existing protocols, the videos in SumMe and TVsum are separated into 80\% for training and the rest 20\% for testing. For simplicity, the validation is operated on the training set. In the training process, the human-created summaries in SumMe and TVsum are modified to the frame-level importance scores, which is taken as the supervision information for AVRN optimization. Practically, the summarization performance varies among different videos. To make the results more convincing, random training/test splits are carried out for 5 times, and the average performance is taken as the final results. 

Besides, the OVP \cite{DBLP:journals/prl/AvilaLLA11} and YouTube \cite{DBLP:journals/prl/AvilaLLA11} datasets are employed in the experiment part to augment the training set. They are composed of 89 videos with human-created summaries.

\subsubsection{Feature Extraction}
For visual information, the GoogLeNet \cite{DBLP:conf/cvpr/SzegedyLJSRAEVR15} pre-trained on ImageNet\footnote{http://www.image-net.org/} is adopted to extract visual features for video frames, and the 1024-dim feature vector in pool5 layer is taken as the frame representation. Particularly, considering that neighboring frames are quite similar with each other, the features are extracted for every 15 frames (about 2fps). 

For audio information, the VGGish\cite{hershey2017cnn} pre-trained on Audioset \footnote{https://research.google.com/audioset/index.html} is adopted for audio feature extraction. Specifically, the audio data are temporally separated with the duration of 1 second. Two neighboring segments share the overlap of half a second, so that the length of audio feature can match that of the visual feature. Note that there are two videos in SumMe without audio data, 'Scuba' and 'St Maarten Landing'. Their audio features are padded with zeros.

\subsubsection{Performance Evaluation}

The summary quality is evaluated by measuring the temporal consistency between predicted summary and human-created summary. Precision (P), Recall (R) and F-measure (F) are widely adopted metrics. They are defined as 
\[P = \frac{{\# \left( {summar{y^p} \cap summar{y^h}} \right)}}{{\# summar{y^p}}},\]
\[ R = \frac{{\# \left( {summar{y^p} \cap summar{y^h}} \right)}}{{\# summar{y^h}}},\]
\[F = \frac{{P \cdot R}}{{\left( {P + R} \right)}}.\]
where \({\# summar{y^p}}\) and \({\# summar{y^h}}\) denote the number of frames in the predicted summary and human-created summary, respectively. \({\# \left( {summar{y^p} \cap summar{y^h}} \right)}\) stands for their overlap. Considering that each video contains multiple human-created summaries, the pairwise comparisons are conducted. Following existing protocols, the maximum scores are taken as the results on SumMe. The average scores are taken as the results on TVsum dataset. 

To evaluate the performance comprehensively, the rank-based evaluation metrics are also adopted in this work. They are Kendall's \(\tau\) and Spearman's \(\rho\) \cite{otani2019rethinking}, which measure the correlation coefficients of the generated importance scores and annotated importance scores. The pairwise evaluations are conducted among multiple annotated importance scores, and the average coefficients are taken as the final results on both the SumMe and TVsum datasets.

\subsection{Ablation Studies}
The proposed AVRN is composed of three parts, including the Two-Stream LSTM (TS-LSTM), AudioVisual Fusion LSTM (AVF-LSTM), and the Self-Attention Video Encoder (SAVE). To verify the effectiveness of each part, the ablation studies are conducted, and several baselines are compared, including:

\begin{itemize}
\item Audio LSTM: A bidirectional LSTM is employed to encode the audio feature and generate the video summary, while the visual feature is ignored.

\item Visual LSTM: A bidirectional LSTM is employed to encode the visual feature and generate the video summary, while the audio feature is ignored.

\item TS-LSTM: Only the two-stream LSTM is conducted to capture the temporal dependency among audio and visual features, and generate the video summary.

\item AVF-LSTM: The audio and visual features are fused directly without exploiting their temporal dependencies. 

\item AVRN(w/o SAVE): The global video information is not considered when predicting the video summary.

\item AVRN(single): The bidrectional LSTMs in TS-LSTM and AVF-LSTM are replaced with single LSTM (forward LSTM).
\end{itemize}

Table \ref{Table1} exhibits the results of ablation studies. The Audio LSTM can get satisfactory results just with the audio feature as input, which indicates the audio modality can indeed provide useful information for video summarization. TS-LSTM outperforms the Visual LSTM and Audio LSTM significantly. It proves the necessity to integrate audio and visual information together to boost the performance. AVF-LSTM takes the audio and visual features as input, and fuses them without utilizing the TS-LSTM to capture their temporal dependency. AVRN(w/o SAVE) equals to the combination of TS-LSTM and AVF-LSTM. AVRN(w/o SAVE) outperforms them. It verifies the importance of temporal dependency among audio and visual features, and the necessity to fuse the audio and visual information to achieve the mutual benefit between them. 

The difference between the proposed AVRN and the baseline AVRN(w/o SAVE) lies in that the self-attention video encoder is not included in AVRN(w/o SAVE). It means that only the temporal dependency is captured in AVRN(w/o SAVE), while the global video information is ignored. The better performance of the proposed AVRN indicates the global video information are also very important to the summary quality. Besides, AVRN(single) gets worse results than AVRN that utilizes bidirectional LSTMs to encode and fuse the audiovisual information. It explains the rationality to capture the bidirectional temporal dependencies jointly for video summarization. Overall, the results in Table \ref{Table1} have demonstrated the effectiveness of the proposed AVRN, including the two-stream LSTM, the audiovisual fusion LSTM and the self-attention video encoder.

\subsection{Comparison with state-of-the-arts}

Table \ref{Table2} presents the results of AVRN and traditional approaches. The compared approaches are in various categories. \(k\)-medoids, Delauny and VSUMM are clustering based approaches. They are developed based on \(k\)-medoids, delauny clustering and \(k\)-means, respectively. Particularly, considering that the video frames vary smoothly, the clusters in VSUMM are initialized by segmenting video temporally, so that better results are achieved. It indicates the domain knowledge of video is important for the summarization task. SALF and LiveLight are dictionary learning based approaches, where SALF generates summary by self-reconstructing the video with shots sparsely, and LiveLight conducts an online learning strategy to incrementally select those shots that cannot be represented by current key-shot set. They get better performance than clustering based approaches. It is mainly because they can capture the dependency among frames. CSUV and LSMO are designed based on manual criteria. CSUV selects key-shots according to their interestingness measured by factors of aesthetics, landmarks, faces, persons, objects, \emph{etc.}. Inspired by it, LSMO further constructs the interestingness, representativeness and uniformity models. CSUV and LSMO are supervised approaches, so that they can outperform the unsupervised clustering and dictionary learning based approaches. 

The proposed AVRN maintains the advantages of traditional approaches. It can capture the temporal and global dependency among frames by the LSTM and global video encoder. It can also be optimized under the supervision of human-created summaries. Moreover, AVRN utilizes audio and visual features jointly to predict the summary, which means more information is exploited than traditional approaches just extracting visual features. Therefore, it outperforms the traditional approaches significantly.

\begin{table}[tp]
	\centering
	\caption{The results of AVRN and traditional approaches.}\label{Table2}
	\renewcommand\arraystretch{1.2}
	
	\begin{tabularx}{0.45\textwidth}{p{3cm}<{\centering}|X<{\centering}X<{\centering}}
		\toprule\hline
		
		Datasets &SumMe&TVsum\\ \hline
		\(k\)-medoids \cite{hadi2006video} &0.334&0.288\\
		Delauny \cite{DBLP:journals/jodl/MundurRY06}  &0.315 &0.394  \\
		VSUMM \cite{DBLP:journals/prl/AvilaLLA11} &0.335 &0.391 \\
		SALF \cite{DBLP:conf/cvpr/ElhamifarSV12} &0.378 &0.420  \\
		LiveLight \cite{DBLP:conf/cvpr/ZhaoX14a} &0.384&0.477 \\
		CSUV \cite{DBLP:conf/eccv/GygliGRG14}&0.393	&0.532	\\
		LSMO \cite{gygli2015video} &0.403	&\underline{0.568}	\\
		Summary Transfer \cite{DBLP:conf/cvpr/ZhangCSG16} &\underline{0.409}	&--	\\
		AVRN &\textbf{0.441}	&\textbf{0.597}\\
		\hline\bottomrule
	\end{tabularx}
	
\end{table}

Table \ref{Table3} presents the results of AVRN and other deep learning approaches, in order to show the superiority of AVRN. vsLSTM firstly employs bidirectional LSTM to capture the temporal dependency among frames and summarize the video. dppLSTM extends vsLSTM by utilizing the DPP model to guarantee the diversity of key-shots. However, they are plain sequence models without considering the global video information, so they perform much worse than the proposed AVRN. vsLSTM-att and dppLSTM-att are modified by adding attention models to encode the global video information. By exploiting both the temporal and global video information, vsLSTM-att and dppLSTM-att perform much better than the original ones. A-AVS is also an attention based LSTM model, which performs comarably with vsLSTM-att and dppLSTM-att. It has demonstrated the necessity to integrate global encoder to the sequence model for the video summarization task. That's why the self-attention video encoder is developed in AVRN. Furthermore, the even better performance of AVRN has verified the superiority of utilizing both the audio and visual feature to summarize the video.

\begin{table}[tp]
	\centering
	\caption{The results of AVRN and deep learning approaches.}\label{Table3}
	\renewcommand\arraystretch{1.2}
	
	\begin{tabularx}{0.45\textwidth}{p{3cm}<{\centering}|X<{\centering}X<{\centering}}
		\toprule\hline
		
		Datasets &SumMe&TVsum\\ \hline
		vsLSTM \cite{DBLP:conf/eccv/ZhangCSG16} &0.376 &0.542  \\
		dppLSTM \cite{DBLP:conf/eccv/ZhangCSG16}&0.386 &0.547 \\
		SUM-GAN \cite{DBLP:conf/cvpr/MahasseniLT17}&0.387 &0.508  \\
		SUM-GAN\(_{sup}\) \cite{DBLP:conf/cvpr/MahasseniLT17} &0.417&0.563 \\
		H-RNN \cite{DBLP:conf/mm/ZhaoLL17}&0.421	&0.579	\\
		HSA-RNN \cite{DBLP:conf/cvpr/ZhaoLL18}&0.423	&0.587	\\
		SASUM \cite{DBLP:conf/aaai/WeiNYYYY18}&0.406	&0.539	\\
		SASUM\(_{sup}\) \cite{DBLP:conf/aaai/WeiNYYYY18}&\textbf{0.453}	&0.582	\\
		A-AVS \cite{ji2019video}&0.439	&\underline{0.594}	\\
		DR-DSN \cite{DBLP:conf/aaai/ZhouQX18}&0.414	&0.576	\\
		DR-DSN\(_{sup}\) \cite{DBLP:conf/aaai/ZhouQX18}&0.421	&0.581	\\
		SMIL \cite{lei2018action}&0.412	&0.513	\\
		vsLSTM-att\cite{DBLP:conf/mmm/CasasK19} &0.432 &--\\
		dppLSTM-att\cite{DBLP:conf/mmm/CasasK19} &0.438 &--\\
		WS-HRL \cite{DBLP:conf/mmasia/ChenTWY19} & 0.436 &0.584\\
		AVRN &\underline{0.441}	&\textbf{0.597}\\
		\hline\bottomrule
	\end{tabularx}
	
\end{table}

SUM-GAN proposes to employ the Generative Adversarial Network (GAN) to generate video summary discriminatively, and utilizes the discriminator to achieve the unsupervised learning. SUM-GAN\(_{sup}\) is its supervised version, which outperforms SUM-GAN considerably. SASUM and SASUM\(_{sup}\) take the video captions as the auxiliary information to boost the performance. It is also a multimodal video summarization approach. However, the video captions require much human resources, which limits its applicability. Fortunately, the audio modality matches vision modality naturally in videos. The better results are obtained by AVRN, which shows the advantages of audiovisual fusion than text-visual fusion in the video summarization task. DR-DSN adopts the manual criteria, \emph{i.e.,} representativeness and diversity, to guide the summary generator. The results indicate that manual criteria in traditional approaches can also promote the performance of deep learning approaches.

\begin{table*}[htp]
	\centering
	\caption{The results AVRN and compared approaches in different training data organizations.}\label{Table4}
	\renewcommand\arraystretch{1.2}
	
	\begin{tabularx}{0.9\textwidth}{p{4cm}<{\centering}|X<{\centering}X<{\centering}X<{\centering}|X<{\centering}X<{\centering}X<{\centering}}
		\toprule
		\hline
		Datasets &\multicolumn{3}{c}{SumMe}&\multicolumn{3}{c}{TVsum}\\\hline
		Organizations &Canonical  &Augmented  &Transfer&Canonical  &Augmented  &Transfer  \\\hline
		vsLSTM \cite{DBLP:conf/eccv/ZhangCSG16} &0.376 &0.416&0.407&0.542 & 0.579 &0.569  \\
		dppLSTM \cite{DBLP:conf/eccv/ZhangCSG16} &0.386 &0.429&0.418&0.547 &0.596 &\underline{0.587}  \\
		SUM-GAN \cite{DBLP:conf/cvpr/MahasseniLT17}&0.387 &0.417 &--&0.508&0.589&-- \\
		
		SUM-GAN\(_{sup}\) \cite{DBLP:conf/cvpr/MahasseniLT17} &0.417&0.436 &--&0.563&0.612&--\\
		H-RNN \cite{DBLP:conf/mm/ZhaoLL17}&0.421	&0.438&-- &0.579 &\underline{0.619}&--	\\
		HSA-RNN \cite{DBLP:conf/cvpr/ZhaoLL18}&0.423 &0.421&--	&0.587 &0.598&--	\\	
		DR-DSN \cite{DBLP:conf/aaai/ZhouQX18}&0.414 &0.428&0.424	&0.576 &0.584&0.578	\\
		DR-DSN\(_{sup}\) \cite{DBLP:conf/aaai/ZhouQX18}&0.421 &\underline{0.439}&\underline{0.426}	&0.581 &0.598&\textbf{0.589}	\\	
		VASNet \cite{fajtl2018summarizing}&0.424 &0.425&0.419	&0.589 &0.585&0.547	\\	
		re-SEQ2SEQ \cite{DBLP:conf/eccv/ZhangGS18} &\underline{0.425}&\textbf{0.449}&--&\textbf{0.603}&\textbf{0.639}&--\\
		AVRN &\textbf{0.441}	&\textbf{0.449}	&\textbf{0.432}	&\underline{0.597}	&0.605	&\underline{0.587}\\
		\hline\bottomrule
	\end{tabularx}
	
\end{table*}

\begin{figure*}[t]
	\centering
	\includegraphics[width=0.98\textwidth]{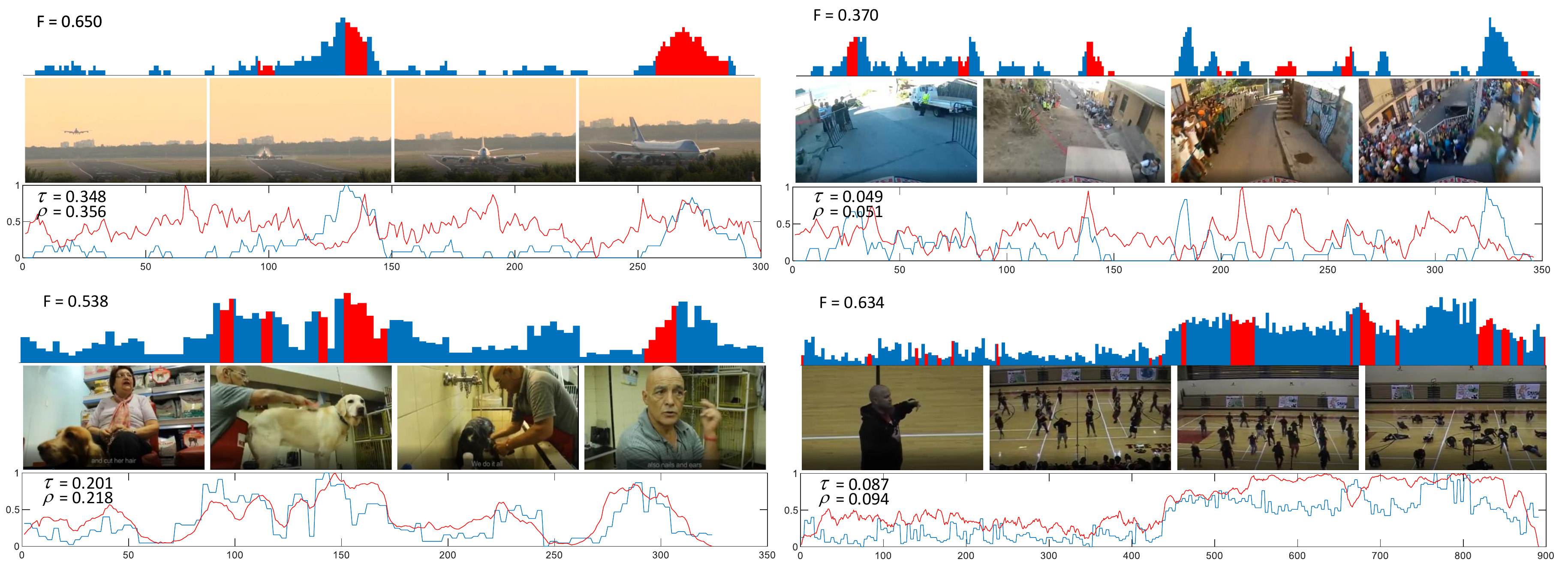}
	\caption{Example summaries generated by AVRN. The above and below samples are from SumMe and TVsum, respectively. The blue curves and red curves below each sample are frame-level importance scores annotated by human and predicted by AVRN. The red histograms above each sample indicate the generated summary. }\label{fig3}
\end{figure*}

H-RNN and HSA-RNN are hierarchical structures of LSTMs. They outperform those approaches with plain LSTM structures, such as vsLSTM and dppLSTM. It is mainly due to the better non-linear fitting ability of the hierarchical structure. Our AVRN also follows the hierarchical structure of LSTM, where the first layer is the two-stream LSTM and the second layer is the audiovisual fusion LSTM. They encode and fuse the audiovisual information hierarchically, so better performance are achieved than plain LSTMs. Furthermore, AVRN also surpasses H-RNN and HSA-RNN. It is mainly because AVRN extracts the multimodal features for video summarization, while H-RNN and HSA-RNN just extract visual features. AVRN can better understand the video content and structure with audiovisual information, so that better results are obtained than H-RNN and HSA-RNN.

In Table \ref{Table4}, different settings of training data are conducted to further analyze the results on SumMe and TVsum datasets. They are canonical, augmented and transfer, which are described as follows:

\begin{itemize}
	\item Canonical: SumMe and TVsum datasets are trained and tested individually. The training/test splits are fixed as 80\% and 20\%.
	\item Augmented: When training on SumMe dataset, the videos in TVsum, OVP and YouTube are employed to augment the training set. Similar strategies are also conducted on the TVsum dataset.
	\item Transfer: The videos in TVsum, OVP and YouTube are utilized to train the SumMe dataset. Similarly, the videos in SumMe, OVP and YouTube are utilized to train the TVsum dataset.
\end{itemize}

From Table \ref{Table4}, we can see that most of the approaches get better results under the augmented setting. This phenomenon indicates that the training data are not enough for most approaches, which leads to the overfitting problem. One effective way to address this problem is to provide more information for the summary generator. DR-DSN exploits the summary properties to reward the summary generator, so better results are obtained than the plain LSTMs, including vsLSTM and dppLSTM. VASNet also adopts an attention model to select key-shots according to the encoded video information\footnote{The results of VASNet are produced by modifying the released source code to the same experimental setting in this paper.}. re-SEQ2SEQ develops a retrospective encoder to keep the consistency of the semantics between video and summary. Different from them, the proposed AVRN introduces the audio features as auxiliary information. In this case, AVRN outperforms most of the approaches just utilizing visual feature for video summarization.

\subsection{Evaluation on rank-based metrics}
\begin{table*}[htp]
	\centering
	\caption{The results of rank-based evaluation (Kendall's \(\tau\) and Spearman's \(\rho\)).}\label{Table5}
	\renewcommand\arraystretch{1.2}
	
	\begin{tabularx}{0.9\textwidth}{p{4cm}<{\centering}|X<{\centering}X<{\centering}|X<{\centering}X<{\centering}}
		\toprule
		\hline
		Datasets &\multicolumn{2}{c}{SumMe}&\multicolumn{2}{c}{TVsum}\\ \hline
		Metrics &Kendall's \(\tau\)  & Spearman's \(\rho\) &Kendall's \(\tau\)  & Spearman's \(\rho\)  \\ \hline
		Random selection  &0.000 &0.000&0.000 &0.000  \\
		dppLSTM \cite{DBLP:conf/eccv/ZhangCSG16}  &-- &--&0.042&0.055   \\
		
		DR-DSN \cite{DBLP:conf/aaai/ZhouQX18} &0.047&0.048&0.020&0.026 \\
		HSA-RNN \cite{DBLP:conf/cvpr/ZhaoLL18}&0.064&0.066&0.082&0.088\\
		AVRN &0.073	&0.074	&0.096	&0.104\\
		Human &0.205&0.213&0.177&0.204\\
		\hline\bottomrule
	\end{tabularx}
	
\end{table*}
Precision, Recall and F-measure quantify the summary quality by measuring the temporal consistency between generated summary and human-created summary. They neglects more fine-grained human preference of video shots. To address this problem, the rank-based metrics, Kendall's \(\tau\)  and Spearman's \(\rho\), are employed in this part to provide comprehensive evaluation of summary quality. They measure the correlation between the predicted probability curve and human-annotated importance curve. 

Table \ref{Table5} presents the results evaluated on rank-based metrics. We can see that the results of the summary generated by random selection is zero. It means there is no correlation between randomly generated summary and human annotation. Besides, considering that each video contains multiple human annotations, the evaluation is also conducted among them via leave-one-out strategy. They get highest scores in Table \ref{Table5}, which shows there are considerable consistency among human annotations. The results meet our expectations.

Some typical RNN-based approaches are compared in Table \ref{Table5}.  dppLSTM is developed based on a plain bidirectional LSTM. DR-DSN conducts representativeness and diversity reward to the summary generator. HSA-RNN constructs the hierarchical structure of LSTM. The proposed AVRN surpasses most of them on Kendall's \(\tau\)  and Spearman's \(\rho\). Besides, Fig. \ref{fig3} displays some exemplar summaries generated by the proposed AVRN. It can be observed from them that AVRN is able to accurately predict the importance scores and effectively summarize the video. The results have demonstrated the superiority of AVRN: 1) The fusion of audio and visual features can provide more information for understanding the video content and structure, so as to benefit the video summarization process. 2) The hierarchical structure of LSTM can enhance the learning ability and further promote the performance. 3) The temporal and global dependency are both very important to the summarization task.

\section{Conclusions}\label{section5}
In this paper, we propose to introduce the audio information for the video summarization task, and develop an AudioVisual Recurrent Network (AVRN) to achieve the fusion of audiovisual features and boost the summarization performance. Specifically, AVRN contains three parts, including the two-stream LSTM, the audiovisual fusion LSTM and the self-attention video encoder. Specifically, the two-stream LSTM can capture the temporal dependency among audio features and video features, respectively. The audiovisual fusion LSTM can exploit the latent consistency between audio and visual information. The self-attention video encoder can capture the global dependency in the whole video stream. The experimental results on SumMe and TVsum have demonstrated that 1) the audiovisual multimodal feature can provide more information for the summarization task than the single visual feature. 2) the hierarchical structure can enhance the learning ability of LSTM. 3) the fusion of audio and visual features and the integration of temporal and global dependencies are both necessary for the video summarization task.

\bibliographystyle{IEEEtran}
\bibliography{IEEEtrans}

\end{document}